\documentclass[%
 reprint,
 amsmath,amssymb,
 aps,
nofootinbib, floatfix]{revtex4-2}

\DeclareMathOperator*{\argmax}{arg\,max}
\DeclareMathOperator*{\argmin}{arg\,min}

\usepackage{graphicx}
\usepackage{dcolumn}
\usepackage{bm}

\usepackage{floatrow}

\usepackage{hyperref}

\usepackage{cleveref}

\usepackage{subfigure}
\begin{document}

\preprint{APS/123-QED}

\title{State Estimation of Power Flows for Smart Grids via Belief Propagation}
\author{Tim Ritmeester}
\affiliation{Physics and Earth Sciences, Jacobs University Bremen, P.O.Box 750561, 28725 Bremen, Germany.}
\author{Hildegard Meyer-Ortmanns}\email[]{h.ortmanns@jacobs-university.de}
\affiliation{Physics and Earth Sciences, Jacobs University Bremen, P.O.Box 750561, 28725 Bremen, Germany.}
\date{June 11, 2020}

\begin{abstract}
Belief propagation is an algorithm that is  known from statistical physics and computer science. It provides an efficient way of calculating marginals that  involve large sums of products which are efficiently rearranged into nested products of sums to approximate the marginals. It allows a reliable  estimation of the state and its variance of power grids  that is needed for the control and forecast of power grid management. At prototypical examples of IEEE-grids we show that belief propagation not only scales linearly with the grid size for the state estimation itself, but also facilitates and accelerates the retrieval of missing data and allows an optimized positioning of measurement units. Based on belief propagation, we give a criterion for  how to assess  whether other  algorithms, using only local information,  are adequate for state estimation for a given grid. We also demonstrate how belief propagation can be utilized for coarse-graining power grids towards representations that reduce the computational effort when the coarse-grained version  is integrated  into a larger grid. It provides a criterion for partitioning power grids into areas in order to minimize the error of flow estimates between different areas.
\end{abstract}

\maketitle


\section{\label{sec:intro}Introduction}

Monitoring the state of the power grid is essential for ensuring the stability of operation. Fluctuations that are not damped in an adequate manner can lead to widespread blackouts \cite{UCTE}\cite{JacobTran}.
Imbalances in supply and demand must be dealt with locally to avoid congestion of power lines. Variation in demand has always existed, but the increase in renewable energy has dramatically increased variations due to larger and more localized variability of wind and solar energy \cite{WindFluct}.

State estimation is the procedure of inferring an  estimate of the state variables of the power grid, as accurate as possible. In general the state variables are the voltage magnitudes and phase angles, active and reactive power flows and active and reactive power injections.
Research in state estimation has a decades long history \cite{StateEst}, but  recently received more attention; the increasing fluctuations caused by renewable energy penetration require more accurate and especially more frequent monitoring of the power grid. The traditional approach (variations of the least squares method \cite{StateEst}) scales poorly with increasing system size, thus it is not well suited for real-time monitoring of large power grids.

Belief Propagation (BP), also known as message passing, is an algorithm known from statistical physics \cite{CavityMethod,Yedidia} and computer science, artificial intelligence and information science (for a review see, for example,  \cite{Yedidia2}). In statistical physics it is in particular used in the context of replica symmetry breaking, spin glasses, random K-SAT problems \footnote{In K-SAT problems, random variables have to satisfy a set of logical constraints, called clauses, with exactly $K$ variables.}, or error correcting codes. For a review see for example \cite{CavityMethod}.

Recently, BP has been discussed  for applications to power grid state estimation \cite{ExtendedDCState,BPStateEst,BPObserve}.In \cite{ExtendedDCState}, DC-state estimation, including bus voltage angles and  magnitudes as state variables, was treated  with BP, based on heuristic criteria for including angle variables, and illustrated with the IEEE-14 bus system.  The emphasis there was on establishing BP as an efficient method for state estimation. The applicability of BP to the level of the distribution grid has been demonstrated in \cite{BPStateEst} in order to bridge the gap between the traditional energy management system and the micro-grid customer level. As analyzed in \cite{BPObserve}, BP is also suited for a so-called observability analysis of large-scale power grids. The goal of such an analysis is to decide whether a given set of measurements is sufficient for estimating the state and if not, to identify islands for which the flows along all lines can be calculated from available measurements.

The BP method has two established main advantages over traditional algorithms. It is fast even for large networks, as the computation time scales linearly in the system size, and it is robust against extremely large differences in input parameters. Its robustness avoids convergence issues associated with the traditional approach (\cite{LSConverge}). In addition, along with an estimate of the state of the system, BP yields the expected square error on the estimate, so that the validity of the estimate can be directly assessed.

These properties have been leveraged for two different goals. The speed of the BP-based algorithm makes it suitable for real-time state estimation \cite{ExtendedDCState,BPStateEst,GraphState}. The robustness allows to avoid the need for running a separate computation to ensure that the whole system is observable, as is needed in the least squares approach. Furthermore, the algorithm will return estimates of all variables that can be determined by the measurements, even if a subset of variables
is not accessible. This way the amount of pseudo-measurements (estimates based on historical data and forecasts) can be minimized if they are needed to make the system observable \cite{BPObserve}.
Both of these applications concern real-time applications to  power grids, they are relevant mostly from an operational point of view.

In this paper we propose  Belief Propagation  for an analysis of the network from the viewpoint of design. Our first contribution consists of analyzing how the amount of missing data as well as the placement of measurement devices together with the network topology affect the retrievability of a current network state, based on incomplete data.

The second contribution consists of proposing a criterion for coarse-graining the network into a partition such that the flow between areas of the partition is most accurately estimated. The reason is that BP allows  to retrieve flows between different areas, even if not all details inside the areas can be retrieved. At the same time, the coarse-grained version of the power flow distribution reduces the size of the data input when integrated into a larger grid.

In this paper, we restrict our state variables to power flows, which is sufficient for tree graphs in the absence of phase angle measurements and a good approximation if loops are included, as we shall see. To completely characterize the state of the power grid if the topology includes small loops, angle variables must be included in the DC approximation of the grid. An accurate inclusion of angle variables in BP is possible as well, but more subtle, here it will be only briefly addressed and in some detail in subsequent work.

The paper is organized as follows. In section \ref{Sec: StateEst} we relate the state estimation of power flow to Bayes' theorem that is applied to optimize the estimation by utilizing known input on measurements and their error. In section \ref{sec: BP} we summarize BP, its accuracy, computational feasibility and convergence properties. Our results are presented in section \ref{sec: Stat} and \ref{sec: Coarse}, section \ref{sec: Stat} concerning the retrieval of missing data, an optimal placement of measurement units and an assessment of the achievable accuracy of BP. Section \ref{sec: Coarse} deals with coarse-graining the power grid, analyzing partitions with respect to the accuracy of power flow estimation, section \ref{sec: Sum} gives the conclusions with some outlook.

\section{State estimation in terms of power flow}
\label{Sec: StateEst}

\begin{figure}
    \includegraphics[width=\textwidth]{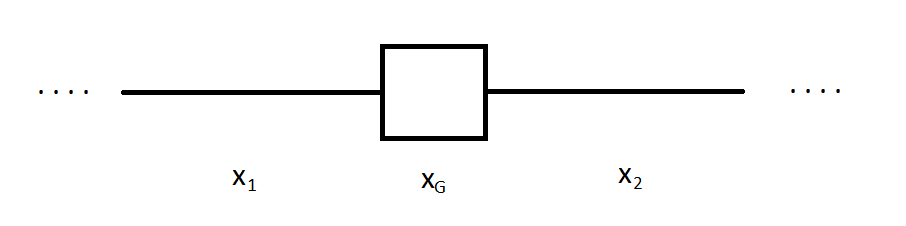}
    \caption{Basic building block of a transmission grid, consisting of two transmission lines and a network bus.}
    \label{fig:simplestatest}
\end{figure}

As a description of the power transmission grid we consider a graph of nodes, representing network buses, and of edges, representing transmission lines. We consider a so-called single-phase equivalent diagram \cite{PowerSystems}. In reality most transmission lines carry three separate conductors. The buses are conductors to which devices, such as conventional power generators or renewable power devices, are connected. The buses are also connected to loads and other parts of the grid, such as a lower-voltage distribution grid. For our purposes, we will collectively refer to power flow from these elements to the network bus as \textit{power injections}, which may be negative and represent consumption.
\\
The transmission grid transmits alternating current. The voltage at any bus is proportional to $\sin(\omega t + \theta_{\mathrm{bus}})$, where $\omega$ is a frequency that is assumed to be constant across the whole network (equal to 50 Hz in continental Europe), and $\theta_{\mathrm{bus}}$ is a bus-specific phase angle. Power flow through high voltage transmission networks is well approximated by a linearization, called the \textit{DC approximation} \cite{DCAccuracy2}, where power flow between two buses is proportional to the  difference in their phase angles, and the proportionality is transmission-line specific but assumed to be known.
\\
Recently, direct measurement of phase angles has become possible by synchronizing measurements to the GPS clock. These devices are called Phasor Measurement Units (PMU    s). However, to date PMUs remain expensive and relatively sparse, they are on the order of 10-100 PMUs across continental Europe \cite{PMUEur}.
\\
In the absence of direct angle measurements, (DC-approximated) power flow in tree networks can be described purely in terms of power conservation. Likewise, if measurements describe flows between areas instead of flows through specific transmission lines (section \ref{sec: Coarse}), angles cannot be specified. For general transmission networks, including loops, the phase angles impose an extra constraint on the power flow. Integration of phase angles into BP is subtle (unless angle measurements are frequent, so that correlations between angles are of shorter range than the loops). An inclusion of phase angles will be considered in subsequent work. Here we will restrict ourselves to networks constrained only by the conservation of power, to demonstrate the utilization options of BP for general networks. 
\\
Let us consider state estimation for an elementary basic building block of power grids, where power is injected to a network bus, connected by two transmission lines to the rest of the power grid (shown in Fig. \ref{fig:simplestatest}).

We denote the state variables by $x_1$ (power flowing through the left hand line), $x_2$ (power flowing through the right hand line) and $x_G$ (power injection at the network bus), defining flow as going from the left to the right. We want to  estimate  these values as precisely as possible, if direct measurements  $z_1$, $z_2$ and $z_G$ of these variables are available. Rather than using these direct measurements as  estimates, we want to exploit the additional information in the form of power conservation: $x_2 = x_1 + x_G$. A combination of this information with directly accessible measurements should lead to an improved estimate. Thus, to estimate $x_2$,  a straightforward average according to $x_2 \approx \frac{z_2 + (z_1 + z_G)}{2}$ is not optimal, as it includes three measurements, while the direct measurement $z_2$ includes only one source of measurement error. 

An appropriate way to combine the direct measurements with an a priori knowledge on the involved errors is given by Bayes' theorem, which turns the state estimation based on direct measurements into a problem of estimating the a posteriori probability that a certain state has led to the observed measurements. Denoting $\mathbf{x} = (x_1, \dots, x_n)$ and $\mathbf{z} = (z_1, \dots, z_N)$:
\begin{align}
    P(\mathbf{x}|\mathbf{z}) = \frac{P(\mathbf{z} | \mathbf{x}) P(\mathbf{x})}{P(\mathbf{z})}.
\end{align}
This is an equation for the a posteriori belief (estimation) that the variable $\mathbf{X}$ had a certain value $\mathbf{x}$, given the measurements $\mathbf{z}$. The probability for the error-prone measurement $P(\mathbf{z} | \mathbf{x})$ is assumed to be known, $P(\mathbf{z})$  and $P(\mathbf{x})$ are known a priori probabilities. From here on, we assume that measurements are drawn as
\begin{equation}
z_a = f_a(\mathbf{x}) + \xi_a,
\end{equation}
where $f_a$ is the measurement function (i.e., the observable that is supposed to be measured), and $\xi_a$ is a random variable that represents the measurement error. The distribution of the error $\xi_a$ is assumed to be known. Here we will assume that it is a Gaussian random variable with $\langle \xi_a \rangle = 0, \langle \xi_a^2 \rangle = \sigma_a^2 $ (the 'measurement variance'), assumed to be known. Given a (locally) uniform prior over values satisfying the conservation of power, Bayes' theorem leads to:
\begin{eqnarray}
    \label{ExplicitBayes}
    P(\mathbf{x}|\mathbf{z}) &\propto &\,   \delta(\mathrm{Conservation}\,\mathrm{of}\, \mathrm{power}) \nonumber \\
    &\times & \prod_a \exp(- (f_a(\mathbf{x}) - z_a)^2/2\sigma_a^2),
\end{eqnarray}
where $\delta(\mathrm{Conservation}\,\mathrm{of}\, \mathrm{power})$ corresponds to $P(\mathbf{x})$ and the second product to the known $P(\mathbf{z} | \mathbf{x})$.
The state estimation problem is then to extract useful information out of this type of expression. Typically we are interested in
\begin{itemize}
    \item the expectations $\langle x_i \rangle_\mathbf{z}$ (which are our best estimates for the state variables), and optionally the expected square error $\langle x_i^2 \rangle_\mathbf{z} - \langle x_i \rangle_\mathbf{z}^2$ (the 'estimate variance'), where $\langle \dots \rangle_\mathbf{z} \equiv \int \mathrm{d} \mathbf{x} \dots P(\mathbf{x}|\mathbf{z})$, and
    \item the marginal distributions $P(x_i | \mathbf{z})$ (from which in turn the moments can be determined).
\end{itemize}
For the example above, we  calculate the marginal distribution of $x_2$ as:
\begin{equation}
    P(x_2|\mathbf{z}) = \int \mathrm{d} x_1 \mathrm{d} x_G P(\mathbf{x}|\mathbf{z}),
\end{equation}
which in this case gives $P(x_2|\mathbf{z}) \sim N\Big(\sigma^2 \big(\frac{z_2}{\sigma_2^2} + \frac{z_1 + z_G}{\sigma_1^2 + \sigma_G^2}\big)\, , \, \sigma^2 \Big)$, where $\sigma^2 = 1/\big(\frac{1}{\sigma_2^2} + \frac{1}{\sigma_1^2 + \sigma_G^2} \big)$. This should be interpreted in the context of Bayes' theorem as a distribution of belief, assigned to $X_2$ having a certain value $x_2$. Our expectation of $x_2$ is thus equal to the mean of this Gaussian. The variance of $P(x_2|\mathbf{z})$ denotes the uncertainty we assign to this estimate.
\\
Note that $\langle x_2^2 \rangle_\mathbf{z} - \langle x_2 \rangle^2_\mathbf{z} = \sigma^2$ is actually independent of $\mathbf{z}$. This is in general the case for linear measurement functions and Gaussian measurement error, in which case Eq.~\ref{ExplicitBayes} is a multivariate normal distribution (in the power flow variables) with mean dependent on $\mathbf{z}$, and covariance matrix independent of $\mathbf{z}$.  The covariance matrix responds linearly to a multiplication of all measurement variances by the same constant: The distribution given by Eq.~\ref{ExplicitBayes} is invariant under changing all variables $x_i \rightarrow x_i' = \lambda x_i$, while simultaneously changing $z_a
\rightarrow \lambda z_a$ and $\sigma^2_a \rightarrow \sigma'^2_a = \lambda^2 \sigma^2_a$ for all factors, where $\lambda$ is independent of $i$ and $a$. Thus multiplying measurement variances by $\lambda^2$ changes estimate covariances to $\langle x_i x_j \rangle - \langle x_i \rangle \langle x_j \rangle \rightarrow \langle x'_i x'_j \rangle - \langle x'_i \rangle \langle x'_j \rangle = \lambda^2 \big( \langle x_i x_j \rangle - \langle x_i \rangle \langle x_j \rangle \big) $, which will be used later.

\section{Belief Propagation} \label{sec: BP}
As we have seen at the elementary example of the last section, the goal is  to calculate marginals, which in general involve a large number of summations or integrations. This is the place where BP comes in, as it provides a rather efficient method to approximate the sums of products as nested products of sums.
BP approximates distributions that factorize into a large number of factors:
\begin{equation}
    P(\mathbf{x}) = \prod_a P_a(\mathbf{x}).
\end{equation}
From an efficiency point of view, the $P_a$ are required to depend  only on a small subset of all variables. Such a distribution can be represented graphically in terms of a \textit{factor graph}. This is a graph with two kinds of nodes, \textit{variables} and \textit{factors}. The factor nodes represent factors of the probability distribution, each being a function of a few variables. A graph is assigned by connecting a factor with all variables it depends on via drawing a line.
\\
To simplify the distribution dictated by Bayes' formula (Eq. \ref{ExplicitBayes}), we will explicitly integrate over all variables representing power injection, thus eliminating the delta function, so that we are left with variables representing only power flow along lines. For the state estimation problem, the factors $P_a$ each represent the probability of a measurement. We will consider injection and power flow measurements.

\begin{figure}
    	\includegraphics[width=0.6\textwidth]{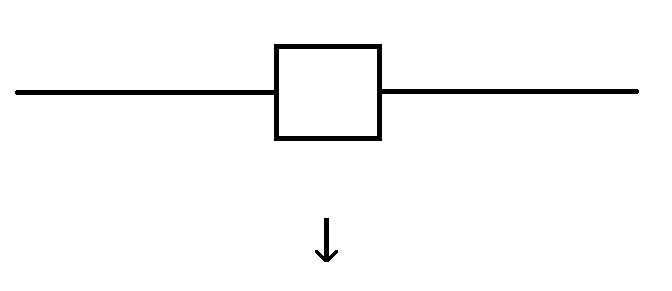}
    	\\
    	\includegraphics[trim= 1600 650 1300 50]{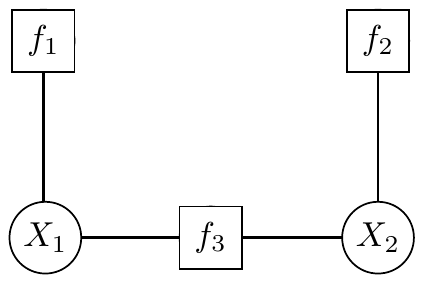}
    	\caption{The transmission grid section from Fig. \ref{fig:simplestatest} transformed to a factor graph.}
    	\label{fig:simplestatestfactorgraph}
\end{figure}

The simple state estimation problem we considered above can then be represented as in Fig. \ref{fig:simplestatestfactorgraph}.

$X_1$ and $X_2$ represent the power flow variables through transmission lines $1$ and $2$ that can take values $x_1$ and $x_2$, respectively.
$f_1$ and $f_2$ are measurements of the power flows $X_1$ and $X_2$, respectively, while $f_3$ represents the measurement of the power injection at the bus connected to the two lines.
For general electricity networks we will use an analogous notation ($f$ for measurements of variables according to $z_a=f_a(\mathbf{x})+\xi_a$, $X$ for variables). We define $\mathcal{N}(Y)$ as the set of neighbours of $Y$ on the factor graph; if $Y$ is a variable, $\mathcal{N}$  returns a factor,  if $Y$ represents a factor, it returns a variable. For the factor graph shown in Fig. \ref{fig:simplestatestfactorgraph} this would mean, for example, that $\mathcal{N}(X_1) = \{f_1, f_3\}$ and $\mathcal{N}(f_3) = \{X_1, X_2\}$.

In general, BP calculates an approximation to the original problem, under certain conditions it gives exact results, as we discuss below. It operates on a space of \textit{pseudomarginals} \cite{MarginalPolytope}. These are sets of `beliefs' $\{b_i(x_i): i \in X\}$ and $\{b_a(\mathbf{x}_a)\: : \: a \in f)\}$ (where X is the set of variables, f is the set of factors and $\mathbf{x}_a \equiv (x_i: X_i \in \mathcal{N}(f_a))$  ) that are meant to approximate the true marginals ($P(x_i|\bm{z})$ and $P(\bm{x}_a|\bm{z})$) and satisfy $\sum_{\mathbf{x}_a \setminus x_i } b_a (\mathbf{x}_a) = b_i(x_i)$ for all $a$, $b_a(\mathbf{x}_a) \geq 0$ for all $a$ and $\mathbf{x}_a$,  and $\sum_{x_i} b_i(x_i) = 1$ for all $i$. For these pseudomarginals, given any variable $X_j$, all variables in $\mathcal{N}^2(X_j) \setminus X_j$ are conditionally independent ($\mathcal{N}^2$ denotes the set of next-to nearest neighbors). BP calculates the marginals in this space which are 'closest' to the true probability distribution, closest in the sense of minimizing the Kullback–Leibler divergence between the distributions \cite{Yedidia}. In the context of statistical physics, this is equivalent to minimizing the free energy in the Bethe approximation \cite{Yedidia,CavityMethod}.

BP then works  by exchanging  messages from variables to factors and from factors to variables until the messages (hopefully, see below) converge to fixed points in the space of pseudomarginals \footnote{At every time step $t$ we have estimates for the marginals $\{b_i^t(x_i)\}$, whose convergence can be tracked to provide a criterion for stopping the iteration. In our case the messages are Gaussian, so that $b_i^t(x_i) = N(\mu_i^t, (\sigma_i^2)_t)$. An appropriate convergence criterion is $\sum_i |\mu_i^{t} - \mu_i^{t-1}| < 10^{-10}$, $\sum_i |(\sigma^{2}_i)_t - (\sigma^{2}_i)_{t-1}| < 10^{-10}$.}. These fixed points are  the marginal probability distributions $P(x_i|\mathbf{z})$ of final interest, here for the state estimation. They are obtained from the following exchange of messages \cite{Yedidia}:
\begin{align}
m_{X_i \rightarrow f_a}(x_i) &= \prod_{f_b \in \mathcal{N}(X_i) \setminus f_a}  m_{f_b \rightarrow X_i} (x_i) \label{messageeq1}\\
m_{f_a \rightarrow X_i}(x_i) &= \int  \Big[  \prod_{x_j \,: \, X_j \in \mathcal{N}(f_a) \setminus X_i}  \mathrm{d} x_j \Big]  \: \Big( P_a(\bm{x}_a)  \nonumber\\
& \quad  \,  \prod_{X_j \in \mathcal{N}(f_a) \setminus X_i}  m_{X_j \rightarrow f_a} (x_j) \Big) \label{messageeq2} \\
b_i(x_i) & \propto   \prod_{f_a \in \mathcal{N}(X_i)}   m_{f_a \rightarrow X_i} (x_i)
\end{align}
with $\mathbf{x}_a$ the set of all $x_j$ such that $X_j \in \mathcal{N}(f_a))$,  the neighborhoods $\mathcal{N}(X_i),\;\mathcal{N}(f_a)$ as defined above,  and
\begin{equation}\label{eq: marg}
b_i(x_i)\rightarrow P(x_i|\mathbf{z}),
\end{equation}
where the beliefs converge to $P(x_i|\mathbf{z})$ in tree networks, and approximately to $P(x_i|\mathbf{z})$ if loops are present. 
We remark  that these Eq.s are equivalent to the transfer matrix and cavity method approaches from statistical physics  \cite{CavityMethod,CVM}. In the context of control theory they are a generalization of dynamic programming and of the Kalman filter algorithm \cite{ContVar,GraphicalModels}.
For the ease of understanding we illustrate  this algorithm at the state estimation of the simple building block represented by Fig. \ref{fig:simplestatestfactorgraph} in  Appendix A.

When using BP for approximate state estimation, some remarks are in order with respect to the accuracy as well as the computational feasibility and convergence.

\subsection{Computational feasibility and convergence of BP}
In the case that the factor graph is a tree, one may start from the leaves and subsequently send the messages to the inner region of the graph to all nodes. When all links in the graph are traversed by messages in both directions, all variables are estimated. In this case no iteration is necessary, the marginal probabilities can be read off. This is what we illustrate in Appendix A. In general and in practice, however,  to calculate the fixed points of the messages in an efficient way (say for a larger factor graph including loops),  some starting values are assigned to the messages (e.g., a Gaussian distribution with zero mean and very large variance). The Eq.s \ref{messageeq1} and \ref{messageeq2} are then iterated (in alternating fashion) until convergence is reached. This procedure has a time complexity linear in the system size. If the factor graph has loops, convergence is not guaranteed, and one might need to resort to a different method of calculating fixed points \cite{Damping} \cite{HAK}. For the problems considered in this paper we report convergence in all cases we investigated.

If factors depend on a large number of variables, computing the messages in Eq. \ref{messageeq2} involves the integration over a large-dimensional space, in this case BP may lose its computational advantage. In our case we assume that the measurements are linear functions of the variables with a Gaussian error, which means that the messages maintain a Gaussian shape. The integrals can then be evaluated analytically. For a general form of the measurement functions and error distributions, the messages are real-valued functions, so there is a need to find a way to represent them numerically. This complicates the applicability of the algorithm, but several methods have been devised for this purpose, e.g.  \cite{ContVar,TreeEP,Noorshams}.
Since the messages in our case maintain a Gaussian shape, they can be represented and forwarded by just two real numbers each (the mean and the variance).
\subsection{Accuracy of BP}
To access the accuracy of BP, it may be instructive  to relate it to familiar approximations such as the Bethe-mean-field approximation of statistical physics. The Bethe approximation keeps track of correlations of a variable and its neighbouring variables, but, conditional on the variable itself, it sets correlations between different neighbours to zero due to the product ansatz for the trial distribution. This is exact if the factor graph is a tree, but loses accuracy if there are (short) loops. More precisely, for a tree topology both of the original grid and the factor graph, BP is exact in the sense that the Kullback-Leibler divergence between the distribution of beliefs and the true marginal probability distribution goes to zero \cite{Yedidia}. It is almost exact, if the length of the loops is of the order of  the system size, so that the graph is locally tree-like \cite{CavityMethod}.

Real power grids as well as our test systems contain a number of short loops. In these cases, there is no guarantee that BP converges at all, or that it converges to a probability distribution that approximates the true states within a desired accuracy. In this case it is convenient to first check whether so-called "loopy BP" works, that is, whether ignoring the loops leads nonetheless to a fast convergence so that the loops have no detrimental effect.  Otherwise, the BP framework can be extended to incorporate more advanced mean-field methods to improve the accuracy either by diagrammatic expansions \cite{LoopCorrections1,LoopCorrections2} or by  cluster methods like the junction-tree method \cite{Yedidia,CVM}, where essentially whole loops are replaced by  single variable nodes and factor nodes that depend on the variables and factors inside the loops. In principle, this amounts to an exact incorporation of loops, but it may slow down the algorithm considerably.

\subsection{Belief propagation for IEEE-grids in comparison to the least squares method}
{\bf The least squares method.} Until recently, the least squares method was the standard approach to power grids' state estimation.
For a multivariate Gaussian, $\langle x_i \rangle_P = \hat{\mathbf{x}}_i$, where $\hat{\mathbf{x}} = \argmax_{\mathbf{x}} P(\mathbf{x})$. For the distribution of Eq.~\ref{ExplicitBayes} this can be determined by $\argmax_{\mathbf{x}} P(\mathbf{x}|\mathbf{z}) = \argmin_{\mathbf{x}}  \sum_a (f_a(\mathbf{x}) - z_a)^2/2\sigma_a^2$. Least squares algorithms perform this minimization with the aim of finding $\{ \langle x_i \rangle_P\}$. In practice, such algorithms can find the minimum by performing a matrix inversion.
For linear $\{f_a\}$ we can define a symmetric matrix $\mathbf{A}$  and vectors $\mathbf{B}$ and $\mathbf{C}$ such that  $ \sum_a (f_a(\mathbf{x}) - z_a)^2/2\sigma_a^2 = \mathbf{x}^T \mathbf{A} \mathbf{x} + \mathbf{B} \cdot \mathbf{x} + \mathbf{C}$, which implies $\hat{\mathbf{x}} = -\frac{1}{2} \mathbf{A}^{-1} \mathbf{B}$. Computing matrix inverses algebraically is computationally intensive, but as $\mathbf{A}$ is usually sparse (and is symmetric),  numerical algorithms for sparse matrices are usually more effective \cite{StateEst}. \\
Furthermore, if the system is under-determined (`unobservable'), the matrix  $\mathbf{A}$ in the least squares method is singular and thus not invertible, in which case  numerical matrix inversion algorithms might experience convergence issues. Hence an observability analysis is required prior to running the least squares algorithm. If it turns out that the system is not observable, pseudo-measurements should be added. We use this method only for comparison with our BP results.\\

{\bf IEEE-grids.} We will perform our analysis on the IEEE-118 and IEEE-300 benchmark systems \cite{IEEEBenchMark}. IEEE-grids have been proposed as synthetic electric grid test cases to be statistically and functionally similar to actual electric grids without revealing  any confidential critical information about the energy infrastructure. In particular, a clustering technique is employed which ensures that the synthetic substations meet realistic proportions of load and generation and other constraints. Also synthetic line topologies are chosen with realistic parameters of structural statistics, and connectivity.\\
This is in  contrast to synthetic power grids, often used in the physics community, in which production and consumption are set to $\pm1$, as if they were spin values, and where values for the susceptance and inertia  are chosen homogeneously. It is well known that the grid topology alone does not determine the dynamics, but the assignment of inhomogeneous values for production, consumption, susceptance and inertia can have a strong impact on the grid stability and synchronization patterns. In appendix B we illustrate the inhomogeneous values with some data from the IEEE-14 grid.
 \\
The IEEE-118 network has 118 buses, 186 lines and 65 loops, the IEEE-300 network has 300 buses, 411 lines, and 111 loops. The data of the IEEE-grids contain  phase angles, originally determined from flow measurements, consumption and production values at buses, and characteristics for the transmission lines like susceptance values. From these phase angles and susceptance values we first calculate within the DC approximation  the power flow along the lines and the production and consumption values at the buses as input data for our BP. (Since these values are obtained within the DC approximation, the production and consumption values differ from the corresponding data in the IEEE-grid.) In case that the power grid would have a tree-structure, the restriction to flow values as the only variables in the factor graph would correspond to an exact treatment. If the power grids like IEEE-grids contain loops,
constraints to angle variables around loops require in principle an inclusion of angle variables as variables in the factor graph, inducing additional loops in the factor graph. How to treat these loops will be the topic of a forthcoming paper.

However, ignoring angle variables in spite of the fact that the IEEE-grids contain  loops, has  no detrimental effect and is not needed as long as phase angle measurements are rarely accessible so that we do not want to estimate phase angles themselves.  On the IEEE-300 network, we find that for measurement variances of $10^{-4}$ for both power flow and power injection measurements, the difference in total squared error with the least squares results (that include phase angles) is on the order of $10^{-6}$.

\begin{figure}[ht]
    \subfigure{
    \includegraphics[width = 0.8 \textwidth]{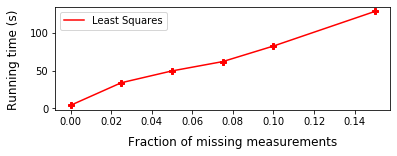}
     \llap{\parbox[b]{17.0cm}{\textbf{\vspace{-3.8cm}\hspace{2cm}(a)}\\\rule{0ex}{5.2cm} }}
    }
    \subfigure{
    \includegraphics[width = 0.8 \textwidth]{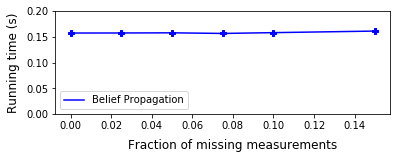}
     \llap{\parbox[b]{17.0cm}{\textbf{\vspace{-3.8cm}\hspace{2cm}(b)}\\\rule{0ex}{5.2cm} }}
    }
    \caption{Performance of a) the least squares method and b) Belief Propagation (for state estimation on the IEEE-118 network), as a function of the  fraction of nodes that do not have direct measurements (selected uniformly according to the description below). Each data point is an average over 50 runs of the algorithms.}
    	\label{fig:speedmissing}
\end{figure}

\begin{figure}[ht]
	\includegraphics[width = 0.8 \textwidth]{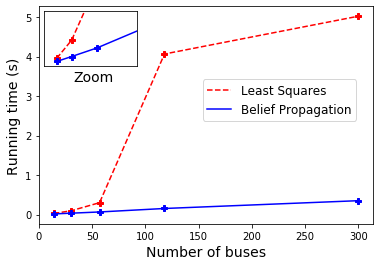}
     	\caption{Performance of the state estimation algorithms as a function of the network size, for the IEEE-14, -30, -57, -118 and -300 networks. Each data point is an average over 50 runs of the algorithm.}
     	\label{fig:speeddifferentnetworks}
\end{figure}

One of our aims is to show how BP can be used to extract {\it statistical} information about electrical grids. Statistical analysis of state estimation problems on large networks is infeasible with conventional methods. In the best case scenario (high precision measurements of all variables), state estimation on one of the power grids we will consider, the `IEEE-300' network, takes ~0.3 seconds to run with BP, compared to ~5 seconds for SciPy's Least Squares implementation \cite{Scipy}\footnote{Both are implemented in Python 3.7 on a 64 bit Windows 10 PC with an Intel i5-2400 Core and a 3.10 GHz CPU.}. The least squares algorithm is sensitive to the numerical values of the input, and the performance very quickly degrades if less or lower precision measurements are considered. Fig. \ref{fig:speedmissing} shows the dramatic increase in computational time in comparison to BP when some fraction of measurements is `missing' (that is, when a fraction of the measurement variances is set to infinity).  In addition, due to the linear scaling of BP, its superiority increases with growing system size \cite{GraphState} \cite{GNState}, as shown in Fig. \ref{fig:speeddifferentnetworks}. For example, Ref.~\cite{BPObserve} uses BP to investigate extremely large systems (up to $7 \times 10^5$ network buses) in a time-scale of seconds. For realistic power systems, the need for using BP for these purposes is thus even more apparent than for our relatively small test grids.

\section{Results of the Statistical Analysis} \label{sec: Stat}
The efficiency of the BP algorithm makes it a promising approach for real-time monitoring of the power grid. Here we show that it also allows  statistical analyses of power grids that were previously inaccessible. We will investigate the following questions:
\begin{itemize}
    \item How does the observability and the need for pseudo-measurements depend on the amount of missing measurements?
    \item To what extent can a strategic measurement placement  improve both of these aspects? Here an extension to phase angles  implemented as variables in the factor graph would be  of particular interest, as an optimal placement of PMUs is an active area of research \cite{PMUOpt1, PMUOpt2, PMUOpt3, PMUOpt4}.
    \item How can we assess whether a generic local algorithm (that includes information only from a local neighborhood) is suited for providing  a detailed analysis of a given grid? Such algorithms are commonly used for optimal PMU placement \cite{PMUOpt1, PMUOpt3, PMUOpt4}. As we shall see, BP (operating locally, but including non-local information) can provide such an observability analysis of the grid and suggest whether a local algorithm will give good estimates for the state variables of the grid.
\end{itemize}

\subsection{Observability, missing data and their retrieval}
For a given network topology, we will generate a large number of 'sample' measurement vectors, and run the state estimation algorithm on those. For every sample, we will  draw measurement values not just randomly, but also vary the locations where measurements are considered as 'missing'.

\begin{figure}
 	\includegraphics[width=0.7\textwidth]{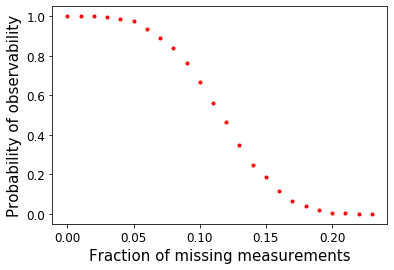}
 	\caption{Probability of the IEEE-300 network being observable as a function of the percentage of missing measurements. Each data point is an average over $5 \cdot 10^3$ samples. The standard error is too small to be visible.}
 	\label{fig: Observability}
 \end{figure}

\begin{figure}
	\includegraphics[width = 0.7\textwidth]{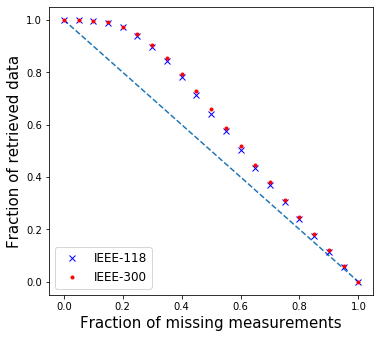}
          \caption{Probability of individual variables of being retrievable as a function of the percentage of missing measurements, for the IEEE-118 and IEEE-300 networks. The dotted line represents the hypothetical case where none of the missing measurements would be retrievable. Each data point is an average over $5 \cdot 10^3$ samples. The standard error is too small to be visible.}
    	\label{fig:PercentRetrieval}
\end{figure}

\begin{figure}
	\includegraphics[width = 0.8 \textwidth]{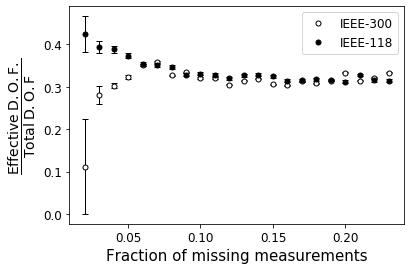}
     	\caption{Number of effective degrees of freedom (D.O.F.) as a function of the fraction of missing measurements. Data points are averages over $5 \cdot 10^3$ runs each. }
     	\label{fig: EffectiveDOF}
\end{figure}

Measurements will be made as before. For each sample, measurements $z_a$ are drawn as $f_a(\mathbf{x}) + \xi_a$, where the measurement error $\xi_a$ is a random variable. If $z_a$ is a measurement of the power flow through line $j$, then $f_a$ is simply the identity map on $x_j$. If it is a measure for power injection, the corresponding function is the sum of all power flowing outwards through connecting transmission lines (as dictated by the conservation of power). If we only want to analyze whether missing measurements would be retrievable (Fig. \ref{fig: Observability}, \ref{fig:PercentRetrieval}, \ref{fig: StrategicC}, \ref{fig: StrategicM}, \ref{fig: ConnectivityCorr}, \ref{fig: NeighbourCorr}, \ref{fig: EffectiveDOF} and \ref{fig:Neighbourhoodretrieval}), the actual distribution from which $\xi_a$ is drawn is irrelevant, as these results only depend on whether a measurement is present at all with an assigned finite variance. Nevertheless, for concreteness we will specify a specific distribution from which the errors are chosen. This will furthermore allow us to take some conclusions about the accuracy of the resulting estimates. In principle, the distribution from which the errors are drawn will depend on any specific situation one wishes to model. Here we will only demonstrate the applicability of the algorithm.\\ \\
{\bf Retrievable variables and observable systems.} We choose $\xi_a$ to be Gaussian random variables with $\langle \xi_a \rangle = 0, \langle \xi_a^2 \rangle = \sigma_a^2$. For testing,  we will formally assign a measurement to every line and every bus in the network.  If for a given sample a measurement is considered as missing, we set its variance $\sigma_a^2 \rightarrow \infty$. For other measurements we uniformly set the variance equal to $1 \times 10^{-4}$, both for flow and injection measurements. If the state estimation algorithm returns a finite variance of the marginal distribution of a variable (Eq. \ref{eq: marg}), the variable is called \textit{retrievable}. If all variables are retrievable, the whole system is \textit{observable}.

We start with the basic question of how many measurements are needed to make the system observable. We first specify the percentage of missing measurements. For every sample, we choose missing flow and injection measurements at random (such that the total number of missing measurements conforms to the pre-specified percentage) and set their variances to infinity. Next we draw measurements according to the measurement variances, and run the state estimation on the resulting measurement values. If this set of measurements is sufficient to determine an estimate of a variable, the state estimation algorithm will return a marginal distribution of that variable with a finite variance.

We present results first for an equal fraction of missing injection and flow measurements.
In Fig. \ref{fig: Observability}, the probability that the system is completely observable is shown for the IEEE-300 network as a function of the fraction of missing measurements. We see that, for the IEEE-300 network, up to $\sim$ 2 \% of missing measurements can be tolerated (for which the network is observable in 99.72\% of the cases). For more missing measurements, the observability is increasingly compromised; if 20 \% of measurements are missing, the probability of observability is reduced to 0.3 \%. One can analyse the same cases  for unequal fractions of missing injection and flow measurements.

Less restrictively and alternatively, we consider retrievability of individual data. Here we consider both power injections and power flows as 'data'. Power injection at a bus is retrievable if it has either a direct measurement assigned, or if all power flows through transmission lines connecting to the bus are retrievable. Even if 20 \% of measurements are missing, on average 97.1 \% of all data are still retrievable. \\
Skewing these fractions towards a lower amount of missing flow measurements increases the amount of retrievable variables, for example, if respectively 20\% and 50\% of flow and injection measurements are missing,
89.0\% of all data can be retrieved, whereas if 35\% of measurements were missing for both flow and injection, only 85.5\% could be retrieved. In the extreme case, if no flow measurements are missing, all data are retrievable, independently of how many injection measurements are missing. Vice-versa, if less injection measurements and more flow measurements are missing, the retrievability gets worse. For example, if 50\% of flow measurements and 20\% of injection measurements are missing, only 83.5\% of data are retrievable. If 70\% of flow measurements are missing, and no injection measurements are missing, 80.1\% of data are retrievable. \\
 The fraction of data that can be retrieved is very similar between the IEEE-118 network and the IEEE-300 network (Fig. \ref{fig:PercentRetrieval}), in spite of their different size. Thus the degree of retrievability seems to be a generic feature, independent of the network size and detailed topology. This leads us to consider the observability as a function of the retrievability and network size.

{\bf Scaling of the observability with the grid size.} It is of interest what determines the observability $P$ of a grid, that is, the probability that all data can be retrieved, given the retrievability probability $p$ of individual data on injection and flow measurements. Naively, one may expect that $P=p^N$ if $N$ represents the number of injection and flow measurements, treated as independent degrees of freedom. For the two IEEE-grids 300 and 118, which differ by roughly a factor of 3 in the number of injection points and a factor of 2 in the number of flows, we see from Fig.~\ref{fig: EffectiveDOF}  that the number of effective degrees of freedom $N_{eff}$  is almost the same between 5-25$\%$ of missing measurements if $N_{eff}$ is read off from
\begin{equation}
N_{eff}=\frac{\log P }{\log p}
\end{equation}
by measuring P and p for a given number of missing measurements. Stated differently,  the observability of a grid reduces less fast with the grid size than naively expected. What matters is the effective number of degrees of freedom.

\subsection{Optimal placement of measurement units}
\begin{figure}
	\includegraphics[width = 0.8\textwidth]{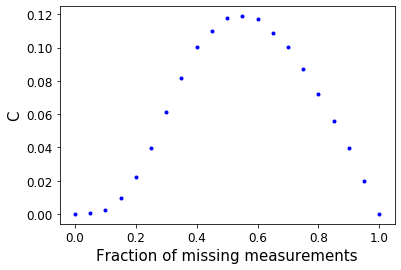}
    	\caption{The correlation C between connectivity and probability of power injection to be retrieved in the IEEE-300 network, as a function of the fraction of missing measurements. Averages over $5 \cdot 10^3$ runs.}
   	 \label{fig: ConnectivityCorr}
\end{figure}

It is interesting to quantify how the retrievability of data depends on the location of missing data along with suggestions where measurement devices should be optimally placed. Based on the fact that future phasor measurement units will be placed at nodes, we focus mainly on missing data at injection points rather than on flows. In principle we could determine case by case what an optimal placement for a given number of measurement units would be, using BP as a very efficient algorithm. Here we will consider the IEEE-300 network and run exactly the same analysis as before, randomly taking out a certain percentage of injection measurements and the same percentage of flow measurements, but keeping track of the correlation between the connectivity of a bus, and the retrievability of the power injection at that bus. We define
\begin{equation} \label{eq: ConnectivityCorr}
C \equiv \langle \frac{\sum_i c_i \delta_i}{300} - \frac{\sum_i c_i \sum_i \delta_i}{300^2}\rangle,
\end{equation}
where $c_i$ is the connectivity of bus $i$ and $\delta_i$ is $0$ if the power injection at bus $i$ is retrievable (in a given sample), and $1$ otherwise. The average is taken over the different samples. The results are shown in
Fig.~\ref{fig: ConnectivityCorr}.  Since for $0 \%$ ($100\%$) missing data, all (no) data can be reconstructed, the correlation at these points equals $0$. The correlation can be seen to be positive for any other percentage of missing measurements, meaning that a large connectivity (degree of a node) is positively correlated to being non-retrievable, or $\delta_i=1$, so a high degree of a node decreases the chance of data being recovered. This is natural, since lacking information for even one outgoing line means that  the power injection at a bus cannot be reconstructed (in the absence of a direct measurement).

\begin{figure}
	\includegraphics[width = 0.7\textwidth]{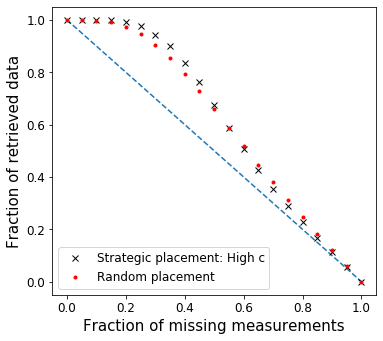}
          \caption{Comparison of the retrievability of variables as a function of the percentage of missing measurements for the IEEE-300 network, where these missing measurements are either randomly selected or preferably at low-connected nodes. Otherwise same as in Fig.~\ref{fig:PercentRetrieval}.}
    	\label{fig: StrategicC}
\end{figure}

This suggests that we can do better than randomly distributing measurement devices across the network buses. Fig. \ref{fig: StrategicC} shows the amount of data that can be retrieved if measurements of buses are deterministically chosen to be missing, starting subsequently from the least connected buses, instead of uniformly at random, which is of less harm for the retrievability of the grid. This strategy does significantly better than a random measurement placement for a small amount of missing data (Fig. \ref{fig: StrategicC}), since missing data at low-connected nodes are more easily retrieved. Therefore it is advantageous to place measurement devices at the higher connected nodes.
Only for a larger amount of missing data (which is not very realistic), the effect diminishes, as we see from Fig. \ref{fig: StrategicC} where for a large amount of missing data, the strategy actually performs slightly worse than a random placement of measurements.

\begin{figure}
	\includegraphics[width = 0.8\textwidth]{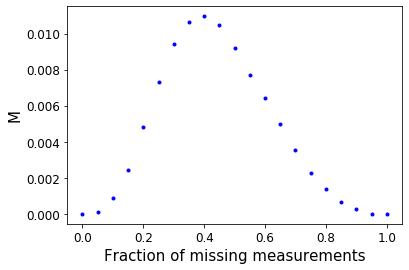}
    	\caption{The correlation $M$ between the amount of missing injection measurements in the neighbourhood of a bus and the retrievability of power injection at the bus, for the IEEE-300 network. Averages are over $5 \cdot 10^3$ runs.}
   	 \label{fig: NeighbourCorr}
\end{figure}

Obviously, the retrievability of data for a bus will not only depend on its connectivity, but also on the retrievability of the buses in its neighborhood.  We therefore consider the correlation
\begin{align} \label{eq: CorrM}
M  \equiv \langle \frac{\sum_i \frac{m_i}{c_i} \delta_i}{300} - \frac{\sum_i \frac{m_i}{c_i} \sum_i \delta_i}{300^2}\rangle \,,
\end{align}
where $m_i$ is the amount of buses in $\mathcal{N}^2(i) \setminus i$ (that is, the buses directly connected to $i$) for which the injection measurement is missing in a given sample. (On a factor graph these buses belong to the next-to-nearest neighbors, thus $\mathcal{N}^2(i)$.) As before, $c_i$ is the connectivity of bus $i$ and $\delta_i$ is $0$ if the power injection is retrievable (in a given sample), and $1$ otherwise. Thus $m_i/c_i$ is the fraction of neighbouring buses of $i$ for which the injection measurement is missing. If the measured correlation is positive it means that a large value of this fraction positively correlates  with the non-retrievability of the node. Note that this value can be large also for a low degree node if the few neighbors have  not  been measured.  Also formally, $\frac{m_i}{c_i}$ is not correlated with $c_i$, or  the value of the correlation $M$ is independent of the correlation $C$ between retrievability and connectivity which we considered before (Eq. \ref{eq: ConnectivityCorr}), since
\begin{align}
\langle \frac{\sum_i \frac{m_i}{c_i}c_i}{300} - \frac{\sum_i \frac{m_i}{c_i} \sum_i c_i}{300^2}\rangle = 0 \,.\text{\footnotemark}
\end{align}
\addtocounter{footnote}{-1} \footnotetext{
This can be shown straightforwardly by using the fact that the connectivity $c_i$ is independent of the sample and that $\langle \frac{m_i}{c_i} \rangle$  is simply the fraction of injection measurements that is missing. Thus it is independent of $i$ if the missing measurements are randomly chosen. } \stepcounter{footnote}
The correlation $M$ (Eq. \ref{eq: CorrM}) is shown in Fig. \ref{fig: NeighbourCorr} as a function of the fraction of missing measurements. $M$ turns out to be always positive, although it is roughly an order of magnitude smaller than $C$ (Fig. \ref{fig: ConnectivityCorr}).

\begin{figure}
	\includegraphics[width = 0.7\textwidth]{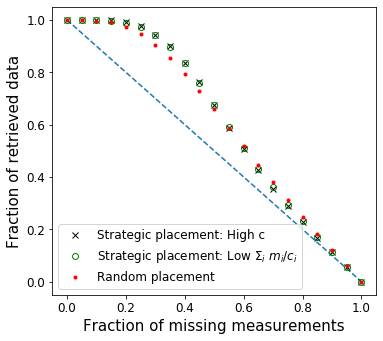}
          \caption{Same as Fig.~\ref{fig: StrategicC} with added data from a strategic placement of missing measurements, located such that $\sum_i \frac{m_i}{c_i}$  is minimized. For further explanations see the text.}
    	\label{fig: StrategicM}
\end{figure}

This observation once more suggests that we can strategically distribute measurements such that more data is retrievable than  for a random placement of measurements. Fig. \ref{fig: StrategicM} shows the amount of data that can be retrieved if measurements at buses are chosen to be missing deterministically, such that $\sum_i \frac{m_i}{c_i}$ is minimized for the given overall fraction of missing measurements (in particular $m_i$ should be zero if $c_i=1$, not to contribute to the non-retrievability). In Fig. \ref{fig: StrategicM} it can be seen that for a reasonably small amount of missing measurements this strategy performs better than random measurement placement. In fact, the performance of this measurement placement is very close to that of the strategic placement which we considered before, where measurements were only missing at low-connected buses as it was less harmful. Empirically (not shown here), the measurement configurations that minimize $\sum_i \frac{m_i}{c_i}$ tend to place the measurements at high-connectivity buses (realized as deleting measurements at nodes for which the contribution to $\sum_i \frac{m_i}{c_i}$ is low). \\
The fact that the configurations provided by these two strategies are very similar might thus explain the observation that their performance is also very similar. As before, we thus conclude that placement of measurement devices at high-connectivity buses is a good heuristic criterion for optimizing the retrievability of power grid data.

\subsection{Assessment of the adequacy of a generic local algorithm by means of BP} 
\begin{figure}
    	\includegraphics[width = 0.8 \textwidth]{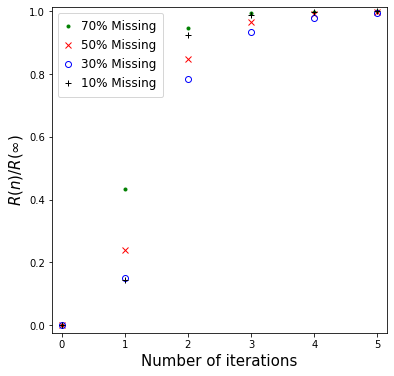}
     	\caption{Dependence of the retrievability on the number of iterations of the BP algorithm for the IEEE-300 network. Data points are averages over $5 \cdot 10^3$ runs each. }
     	\label{fig:Neighbourhoodretrieval}
\end{figure}

Another characteristic of state estimation results from recording the amount of iterations it takes the BP algorithm to return a finite variance on estimates. Each iteration sends a message from the variables to neighboring factors, and then a message from the factors to neighboring variables.
For $n$ iterations of the BP algorithm, for a variable $i$, information from all measurements (factors) in a neighborhood $\mathcal{N}^{2n -1}(i)$ is contained in the estimate for $P(x_i|\mathbf{z})$. Thus the number of BP-updates it takes the algorithm to return a finite expected square error on an estimate is a measure for the size of neighborhood that has to be taken into account for the variable to be retrievable. \\
In Fig. \ref{fig:Neighbourhoodretrieval} we plot a version of this measure for the IEEE-300 network. If $R(n)$ is the amount of variables without direct measurement that can be retrieved after $n$ iterations of the algorithm, the plot shows $\frac{R(n)}{R(\infty)}$ as a function of $n$. The shape of this function depends strongly on the amount of measurements that is missing. For values of around 30 \% missing measurements, the size of the neighborhood that has to be taken into account is largest. On the other hand, if a node is retrievable in spite of 50 or 70\% missing data, the algorithm converges after less iterations, so that less neighborhood of the grid enters the estimation. Of course, in those cases the probability that a node is retrievable is considerably smaller than for less missing measurements.\\

For 30\% missing measurements, the fraction $\frac{R(n)}{R(\infty)}$ is found to be 0.783 for $n = 2$, 0.934 for $n = 3$, $0.979$ for $n = 4$, and $0.993$ for $n = 5$. For 10 \% missing data, these numbers are $0.923$, $0.989$, $0.998$ and $0.9996$. For 30 \% missing data, $R(n) < R(\infty)$ up to $n = 9$. This is remarkably deep into the network, as 46\% of all pairs of buses in the IEEE-300 network are within this distance of each other.

In general, combining measurements accumulates errors on the estimates. Thus, if there is a need for taking a large amount of measurements into account, until the algorithm converges, it may indicate inaccurate predictions. For 30 \% of missing measurements, estimates that can be retrieved only after 4 iterations have, on average, a 5.1 times larger estimate variance (after the algorithm has converged) than those that can be retrieved after a single iteration. If the error is too large to be tolerable, one may want to introduce additional pseudo-measurements. This leads us to two conclusions for designing observability algorithms in general, as suggested by BP.
\begin{itemize}
    \item The achievable accuracy of local algorithms such as \cite{PMUOpt1, PMUOpt3, PMUOpt4} that take into account only a small neighborhood of every node depends on the percentage of missing measurements. This dependence can be estimated from the results above.
    \item If the error on an estimate is too high due to the sparsity of available measurements, one might want to introduce additional pseudo-measurements, even if the network is observable in principle. Many observability algorithms are topological in the sense that only finite and infinite variances are distinguished, but no quantitative information is retrieved. BP may also be used as such a topological algorithm when only finite and infinite variances are distinguished. It will then just serve to perform an observability analysis without any quantitative results. If one does keep track of the value of the variance, as we did here,  one can get a rough estimate of the achievable accuracy if the state estimation should be obtained via a local algorithm. The estimate results from calculating the size of the neighborhood that has to be taken into account in BP for retrieving estimates. As mentioned before,  BP itself operates locally (which is a big advantage), but includes information on non-local neighborhoods.
\end{itemize}

\section{Results for coarse-graining of power grids}
\label{sec: Coarse}
Thus far we have described how state estimation can be used for retrieving precise information on power flow through individual transmission lines. Access to the power flow in coarse-grained versions of power grids, that is, between whole areas or countries, are appreciated in view of embedding these grids into larger ones and reducing the computational efforts. Naturally, there is a trade-off between computational limitations and higher spatial detail, for example, on transmission bottlenecks or exploitation of particular places with desired properties such as good renewable resources \cite{CoarseGrained}. An example would be coarse-grained versions of individual countries, integrated into the European transmission grid.

Moreover, for the response against disturbances and planning of future grid expansions, it is necessary to estimate the flows throughout the network, though very precise spatial information is not needed \cite{CoarseGrained}\cite{DistPartitioning}. Often it is sufficient to know the flows between \textit{areas} of the network. Different ways of partitioning the grids into areas are already available with several benefits, most important of which are a reduction in the computational burden and a decrease in the amount of communication that is needed for real time response \cite{CoarseGrained}\cite{DistPartitioning}. Furthermore, detailed information might simply be unknown - either because of uncertain future grid expansion \cite{CoarseGrained}\cite{TwoPart} or because network operators are unwilling to disclose some information \cite{PowerSystems}.

The total flow between two areas is given by the sum of the flows through the transmission lines connecting those areas, information on the flow through other transmission lines is not needed. The analysis is, however, not completely straightforward since the lack of observability might cause convergence issues. Traditionally, one would first run an observability analysis to find out whether the system is observable. If it is not, one needs to introduce an amount of pseudo-measurements to ensure that it becomes observable. Such pseudo-measurements might, however, corrupt the accuracy of the estimate.

BP has no need for such a separate, additional  observability analysis. It will return an estimate of the flow between areas even if the flow within areas cannot be retrieved. If there are flows between areas that cannot be observed, BP can be used to introduce  a minimum amount of pseudo-measurements to keep those lines observable (along the lines of \cite{BPObserve}).

At present, the areas used for real-time response are largely based on motivations not related to the physics of the grid, but to historically developed zones, or geographical borders, different network operators, or price differences. 
Numerous studies have sought to specify improved criteria for partitioning the power network into areas. Such criteria include geographical closeness  \cite{CoarseGrained,DistPartitioning}, 'electrical distance', based on the connectivity and admittance of transmission lines \cite{ElectricalDist}, the possibility and effect of power transfers between areas \cite{ATC}\cite{PTDF} and
convergence of optimization algorithms \cite{PartOPF}.

We propose as an additional and alternative criterion to choose  a partitioning  such that the   state estimation can be performed with the highest accuracy of the estimated flows when this is of interest for a coarse-grained version of the grid. We will illustrate at the IEEE-14 grid how to determine  the dependence of the precision on the choice of partition, and how to improve this precision.

\subsection{Anti-correlations between estimation errors}

\begin{figure}
    	\includegraphics[width=0.8\textwidth]{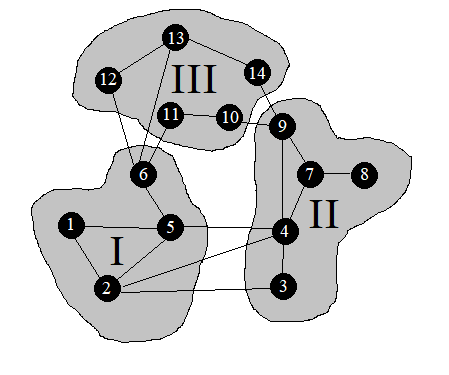}
    	\caption{A partition of the IEEE-14 network based on geographical closeness (`Partition I')}
    	\label{fig:IEEEPartition1}
\end{figure}

Fig. \ref{fig:IEEEPartition1} shows an arbitrary partition of the IEEE-14 network into three areas, whose choice is loosely based on geographical closeness.

The flows between areas can be calculated simply by summing the flows through transmission lines connecting them. Denoting the total flow from region $Y$ to region $Z$ as $Y \rightarrow Z$, the flows (in units of MW) are from $I\rightarrow II: 97.5$, $II\rightarrow III: 10.8$, and $I\rightarrow III: 4.96$.
To determine  errors on these flows, we need to account for  correlations $\langle x_i x_j \rangle_\mathbf{z} - \langle x_i \rangle_\mathbf{z} \langle x_j \rangle_\mathbf{z}$, which in contrast to the estimate variances $\langle x_i^2 \rangle_\mathbf{z} - \langle x_i \rangle_\mathbf{z}^2$ are not determined by the BP algorithm. To calculate the correlations we can use (a basic version of) the linear response algorithm described in \cite{LoopCorrections1}, by utilizing the well known trick:
\begin{align}
P(\mathbf{x}|\mathbf{z}) \propto \,  &\frac{\exp(-(x_j - z_j)^2/2\sigma_j^2)}{\int \mathrm{d} \mathbf{x'} P(\mathbf{x'}|\mathbf{z})} \\
\rightarrow \, & \sigma_j^2 \frac{\partial \langle x_i \rangle_\mathbf{z}}{\partial z_j} = \langle x_i x_j \rangle_\mathbf{z} - \langle x_i \rangle_\mathbf{z} \langle x_j \rangle_\mathbf{z}.
\end{align}
For a given set of measurements, we thus perform a separate run of the BP algorithm for every variable we are interested in, and perturb the direct measurement of the variable slightly. This induces a change in the estimate of the other variables, which  is used to numerically evaluate the derivative. We perform this analysis on the IEEE-14 grid with a measurement variance of $1 \times 10^{-4}$ for both power flow and power injection measurements. We find the covariance matrix in units of MW${}^2$ as presented in table \ref{tab:firstpartitioncov}.

\begin{table}[h!]
\begin{center}
\caption{The covariance matrix for the coarse-grained flows assigned to partition I}
\label{tab:firstpartitioncov}
\bgroup
\def\arraystretch{1.5}
 \begin{tabular}{|c c c c|}
 \hline

  & $\mathrm{I} \rightarrow \mathrm{II}$ & $\mathrm{II} \rightarrow \mathrm{III}$ & $\mathrm{I} \rightarrow \mathrm{III}$ \\[0.5ex]
 \hline

 $\mathrm{I} \rightarrow \mathrm{II}$ & 19.7 $\cdot 10^{-4} $ & - 0.619 $\cdot 10^{-5}$ & 5.67 $\cdot 10^{-5}$ \\
 \hline

 $\mathrm{II} \rightarrow \mathrm{III}$ & - 0.619 $\cdot 10^{-5}$ & 8.14 $\cdot 10^{-4} $& - 10.1 $\cdot 10^{-5} $\\
 \hline
 $\mathrm{I} \rightarrow \mathrm{III}$ & 5.67 $\cdot 10^{-5}$ & - 10.1 $\cdot 10^{-5}$ & 8.73 $\cdot 10^{-4} $\\
 \hline
\end{tabular}
\egroup
\end{center}
\end{table}

Crucially, as shown in section \ref{Sec: StateEst}, this matrix is independent of the actual measurements $\mathbf{z}$. The accuracy of state estimation for a given partitioning can thus be calculated up front, independently of the actual measurements and the actual state of the system. The results for partition I in units of MW are given as $\mathrm{I} \rightarrow \mathrm{II} : 97.5 \pm 0.0444$, $\mathrm{II} \rightarrow \mathrm{III} :10.8 \pm  0.0285$, and $\mathrm{I} \rightarrow \mathrm{III} :4.96 \pm  0.0295$.

First of all, the expected square errors for partition I (the diagonal elements of table \ref{tab:firstpartitioncov}) tend to be lower than those given by just summing the variances of the estimate of the individual transmission lines. The inherent  anti-correlations between different estimates can  significantly reduce the error of the coarse-grained variables. An effect of anti-correlations is also seen if the measurement variances are chosen as $1 \times 10^{-5}$, not displayed.

\subsection{Towards an optimal choice for the partition}
\begin{figure}
    \includegraphics[width=0.8\textwidth]{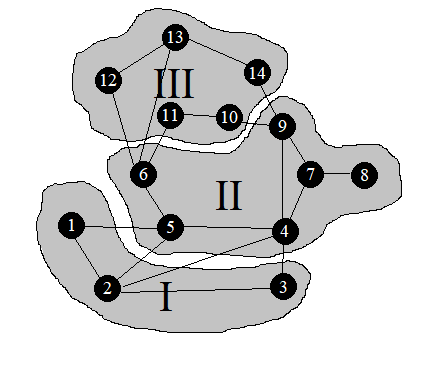}
    \caption{An alternate partition of the IEEE-14 network without transmission lines between I and III (`Partition II').}
    \label{fig:IEEEPartitioning2}
\end{figure}

To illustrate that the choice of partition can in principle be optimized, we consider another partition of the same network, shown in Fig. \ref{fig:IEEEPartitioning2}. This partition gives, for the same measurement variances, the flows in units of MW from $I\rightarrow II: 71.9$, $II\rightarrow III: 15.7$, and $I\rightarrow III: 0$.

\begin{table}[h!]
\begin{center}
\caption{The covariance matrix for flows assigned to partition II}
\label{tab:secondpartitioncov}
\bgroup
\def\arraystretch{1.5}
 \begin{tabular}{|c c c c|}
 \hline

  & $\mathrm{I} \rightarrow \mathrm{II}$ & $\mathrm{II} \rightarrow \mathrm{III}$ & $\mathrm{I} \rightarrow \mathrm{III}$ \\[0.5ex]
 \hline

 $\mathrm{I} \rightarrow \mathrm{II}$ & 11.9 $\cdot 10^{-4}$&  2.34 $\cdot 10^{-4}$ & 0 \\
 \hline

 $\mathrm{II} \rightarrow \mathrm{III}$ & 2.34 $\cdot 10^{-4} $& 14.9 $\cdot 10^{-4}$& 0 \\
 \hline

 $\mathrm{I} \rightarrow \mathrm{III}$ & 0 & 0 & 0 \\
 \hline
\end{tabular}
\egroup
\end{center}
\end{table}

Taking here again the anti-correlations between the errors on the power flow estimates into account  leads to the covariance matrix in units of MW$^2$ of table~\ref{tab:secondpartitioncov} with flows and standard deviations in units of MW given as
$\mathrm{I} \rightarrow \mathrm{II}:  71.9 \pm  0.0345$,
$\mathrm{II} \rightarrow \mathrm{III}: 15.7 \pm 0.0386 $, and
$\mathrm{I} \rightarrow \mathrm{III}: 0$.

To compare the two partitions, we can combine the results for a given partition into a single quantity, the sum of the error squares on the flows. This is given by the trace of the covariance matrices, thus for the two partitions considered it equals, respectively, $36.6 \cdot 10^{-4}$ MW${}^2$ for partition I and $26.8 \cdot 10^{-4}$ MW${}^2$ for partition II. This is a useful metric for finding an optimal  partition of the grid and may be combined with other terms in an objective function, referring to further  criteria \cite{CoarseGrained,DistPartitioning,ElectricalDist,ATC,PTDF,PartOPF}) for selecting a suited partition. One can then employ an appropriate algorithm, such as simulated annealing \cite{SimAn}, to find a partition which optimizes the combined objectives.\\
As long as our variables are restricted to power flows, for a coarse-grained version of the grid we can only estimate the net injection of nodes as the difference between incoming and outgoing power flow. An extension of BP to include production and consumption as variables in the fine-grained version of the grid would allow a coarse-grained version with separate information on production and consumption of the nodes on the coarse-grained level.

\section{Conclusions and Outlook} \label{sec: Sum}
Missing data in power grids are a common source of uncertainty in the management of power grids. At the examples of IEEE-grids we have demonstrated that BP facilitates the retrieval of missing data without the need of a preceding observability analysis. Along with such an analysis, the positioning of measurement units can be optimized by testing the retrieval success as a function of the positioning. This will become relevant for optimal placement of PMUs in the near future. As we have shown, when BP is used as a topological algorithm to distinguish only between infinite and finite achievable variances, its convergence speed is also indicative for the achievable accuracy of state estimation, in particular it indicates whether an algorithm utilizing only local neighborhoods is appropriate for an accurate state estimation.  Due to the ease of data retrieval, independently of missing data inside certain areas, BP can be used for constructing  coarse-grained representations of power production and power flows between entire areas of the grid, together with an estimate of the corresponding variances. Accordingly, the choice of grid partition may be guided by minimizing the uncertainties in flow estimation and combined with other criteria for assigning a reasonable grid partition.

So far our analysis via BP was based  on power conservation, with power flows as the only state variables in the factor graph representation. In forthcoming work we will report on rules of how to implement full DC-approximations, including phase angles as variables in a factor graph. This inclusion leads to extra loops in the factor graphs, while not affecting the number of loops in the original power grid. We will show how to handle these loops for obtaining the true marginals. These loops are small and require only a factor of three more computation time.  Including these loops would leave the results of this paper qualitatively unchanged, the results would slightly change quantitatively. In principle, loops in the power grids themselves can be treated similarly, but their presence may slow down the convergence speed of the algorithm, if the coarse-grained variables and factors representing these loops depend on many variables and factors inside the loops.

We assumed Gaussian errors in the measurement of flow and injection with measurements that are linear functions in the variables. The advantage is that messages in BP which propagate Gaussian distributions can be reduced to two real numbers, their mean and variance, and Gaussians remain Gaussians under subsequent computation of the messages. Other distributions or continuous functions of real numbers should first be discretized, attempts to handle continuous functions as messages are given in \cite{ContVar,TreeEP,Noorshams}. As soon as real functions can be reduced to a small set of real numbers, they are suited  for being propagated as messages.

BP is not restricted to state estimation in the context of power grids, but optimization problems under uncertainty with linear or quadratic cost functions can be  addressed in the same way. This is a promising approach in view of the scalability of BP and will be studied next.

\section*{Acknowledgment} We would like to thank the Bundesministerium f\"ur Bildung und Forschung (BMBF) (grant number 03EK3055D) for financial support.

\appendix
\section{Explicit calculation of the messages}
We illustrate the BP algorithm at the state estimation of the simple building block represented in Fig. \ref{fig:simplestatestfactorgraph}. Starting from
\begin{eqnarray}
P_1(x_1) &=& N(z_1, \sigma_1^2) \nonumber \\
P_2(x_2) &=& N(z_2, \sigma_2^2) \nonumber \\
P_3(x_2 - x_1) &=& N(z_G, \sigma_G^2) , \nonumber
\end{eqnarray}
we can recursively calculate the solution to these equations as:
\begin{eqnarray}
m_{f_1 \rightarrow X_1}(x_1) &=& P_1(x_1) \nonumber \\
m_{f_2 \rightarrow X_2}(x_2) &=& P_2(x_2),\nonumber
\end{eqnarray}
as $f_1$, $f_2$ are the leaves of the tree, so that
\begin{eqnarray}
m_{X_1 \rightarrow f_3}(x_1) &=& m_{f_1 \rightarrow X_1}(x_1) = P_1(x_1) \nonumber\\
m_{X_2 \rightarrow f_3}(x_2) &=& m_{f_2 \rightarrow X_2}(x_2) = P_2(x_2) .\nonumber
\end{eqnarray}
Furthermore,
\begin{eqnarray}
m_{f_3 \rightarrow X_1}(x_1) &=& \int \mathrm{d} x_2 P_3(x_1, x_2) \times m_{X_2 \rightarrow f_3}(x_2) \nonumber \\ &\propto& N(z_2  - z_G, \sigma_2^2 + \sigma_G^2) \\
m_{f_3 \rightarrow X_2}(x_2) &=& \int \mathrm{d} x_1 P_3(x_1, x_2) \times m_{X_1 \rightarrow f_3}(x_1) \nonumber \\ &\propto& N(z_1 + z_G, \sigma_1^2 + \sigma_G^2)
{} \nonumber
\end{eqnarray}
so that
\begin{eqnarray}
P(x_1|\mathbf{z}) &\propto& m_{f_1 \rightarrow X_1} \times m_{f_3 \rightarrow X_1} \nonumber\\
 &\propto& N\Big(\Sigma_1^2(\frac{z_1}{\sigma_1^2} + \frac{z_2 - z_G}{\sigma_2^2 + \sigma_G^2}\big),\, \Sigma_1^2\Big) \nonumber \\
P(x_2|\mathbf{z}) &\propto& m_{f_2 \rightarrow X_2} \times m_{f_3 \rightarrow X_2} \nonumber \\
& \propto& N\Big(\Sigma_2^2 \big(\frac{z_2}{\sigma_2^2} + \frac{z_1 + z_G}{\sigma_1^2 + \sigma_G^2}\big) , \, \Sigma_2^2 \Big),\nonumber
\end{eqnarray}
where
\begin{eqnarray}
&& \Sigma_1^2 \equiv 1/\big(\frac{1}{\sigma_1^2} + \frac{1}{\sigma_2^2 + \sigma_G^2} \big) \nonumber \\
&& \Sigma_2^2 \equiv 1/\big(\frac{1}{\sigma_2^2} + \frac{1}{\sigma_1^2 + \sigma_G^2} \big).\nonumber
\end{eqnarray}

\section{The IEEE test networks}\label{app: IEEE}
\begin{figure}
    \subfigure{
    	\includegraphics[width = 0.8 \textwidth]{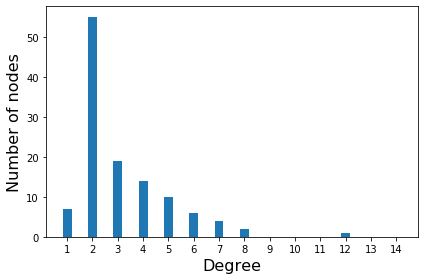}
    	\llap{\parbox[b]{17.0cm}{\textbf{\vspace{-2.85cm}\hspace{2cm}(a)}\\\rule{0ex}{5.2cm} }}
    }\\
    \subfigure{
    	\includegraphics[width = 0.8 \textwidth]{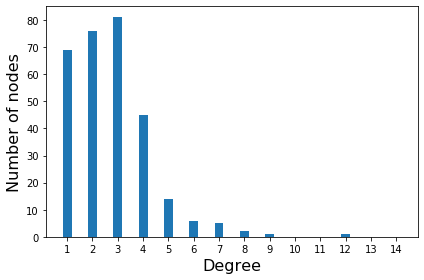}
    	\llap{\parbox[b]{17.0cm}{\textbf{\vspace{-2.85cm}\hspace{2cm}(b)}\\\rule{0ex}{5.2cm} }}
    }
     	\caption{Degree distributions of (a) the IEEE-118 network and (b) the IEEE-300 network.}
     	\label{fig: degree}
\end{figure}

\begin{figure}[H]
	\vspace{20 pt}
	\hspace{225 pt} \includegraphics[trim= 500 750 70 70, scale  = 0.48]{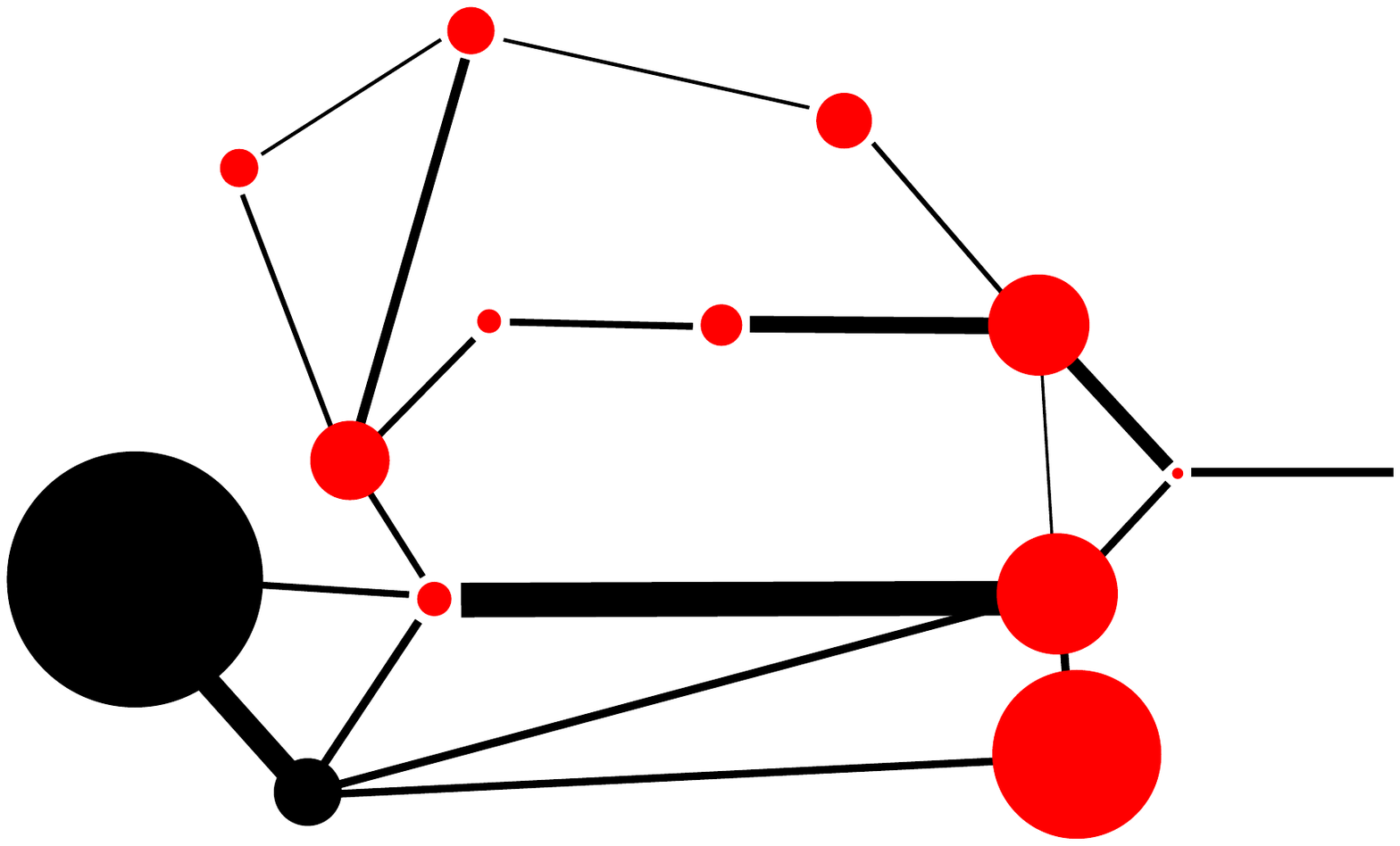}
	\vspace{135 pt}
    	\caption{The IEEE-14 grid with nodes scaled to be proportional to the power injection (nodes shown in red for negative injection), and lines scaled to be proportional to line susceptance to illustrate the inhomogeneity of data.}
    	\label{fig: IEEE14proportional}
\end{figure}

The IEEE benchmark systems are a set of networks that were designed to resemble simplified versions of realistic power networks. De facto, they are the standard in power systems engineering literature to test the applicability of algorithms related to power flow \cite{IEEEsyst}. As the benchmark networks represent real power networks, their structure and parameters are highly heterogenous. In Fig. \ref{fig: degree}, the distribution of the connectivity of the buses in the IEEE-118 and IEEE-300 networks are shown. To further emphasize the heterogeneity of the data, Fig. \ref{fig: IEEE14proportional} shows the IEEE-14 network with the area of the nodes scaled to be proportional to the power injection at the node (with nodes colored in red if the injection is negative, corresponding to consumption), and the width of the lines scaled to their susceptance.

\end{document}